\documentclass[preprint,12pt]{elsarticle}

\usepackage{pgf}

\pgfmathsetseed{\number\pdfrandomseed}

\usepackage{amsfonts}       
\usepackage{amsmath}
\usepackage{amssymb}
\usepackage[ruled,lined]{algorithm2e}
\usepackage{appendix}
\usepackage{booktabs}       
\usepackage{caption}
\usepackage{fancyhdr}       
\usepackage{float}
\usepackage[T1]{fontenc}    
\usepackage{graphicx}       
\usepackage{hyperref}       
\usepackage[utf8]{inputenc} 
\usepackage{lipsum}
\usepackage{microtype}      
\usepackage{mathtools}
\usepackage{nicefrac}       
\usepackage{pgfplots}
\usepackage{pgfplotstable}
\usepackage{tikz}
\usepackage{url}            

\pgfplotsset{compat=newest}

\usetikzlibrary{shapes.geometric, arrows.meta, positioning, calc, patterns}

\tikzstyle{block} = [rectangle, draw, fill=orange!20, drop shadow,
    text width=10em, text centered, rounded corners, minimum height=3em]
\tikzstyle{arrow} = [thick,->,>=stealth, draw=gray]
\tikzstyle{datablock} = [rectangle, draw, fill=teal!20, drop shadow,
    text width=10em, text centered, rounded corners, minimum height=3em]

\graphicspath{{media/}}     

\DeclarePairedDelimiter\floor{\lfloor}{\rfloor}

\newcommand*{\al}{\ensuremath{\alpha}}

\newcommand*{\te}{\ensuremath{\theta}}

\newcommand*{\ph}{\ensuremath{\phi}}

\journal{Knowledge-Based Systems}

\begin{document}

\begin{frontmatter}



\title{Ensemble Distribution Distillation for Self-Supervised Human Activity Recognition}

\author[1]{Matthew Nolan\corref{cor1}}
\ead{matt.nolan@live.com.au}

\author[2]{Lina Yao}
\ead{lina.yao@data61.csiro.au}

\author[3]{Robert Davidson}
\ead{rob.davidson@defence.gov.au}

\cortext[cor1]{Corresponding author}

\affiliation[1]{organization={The University of New South Wales},
                city={Sydney},
                country={Australia}}

\affiliation[2]{organization={The University of New South Wales \& CSIRO},
                city={Sydney},
                country={Australia}}

\affiliation[3]{organization={Defence Science and Technology Group (DSTG)},
                city={Edinburgh},
                state={SA},
                country={Australia}}

\begin{abstract}
    Human Activity Recognition (HAR) has seen significant advancements with the adoption of deep learning techniques, yet challenges remain in terms of data requirements, reliability and robustness. This paper explores a novel application of Ensemble Distribution Distillation (EDD) within a self-supervised learning framework for HAR aimed at overcoming these challenges. By leveraging unlabeled data and a partially supervised training strategy, our approach yields an increase in predictive accuracy, robust estimates of uncertainty, and substantial increases in robustness against adversarial perturbation; thereby significantly improving reliability in real-world scenarios without increasing computational complexity at inference. We demonstrate this with an evaluation on several publicly available datasets. The contributions of this work include the development of a self-supervised EDD framework, an innovative data augmentation technique designed for HAR, and empirical validation of the proposed method’s effectiveness in increasing robustness and reliability.
\end{abstract}


\begin{highlights}
    \item Self-supervised EDD boosts HAR accuracy with minimal labeled data.
    \item HAR-specific augmentations improve robustness and calibration.
    \item Near-ensemble HHAR accuracy with no extra cost and far lower compute/memory.
    \item EDD achieves high robustness against adversarial inputs.
    \item Robust, well-calibrated, separable uncertainty under clean and adversarial inputs.
\end{highlights}


\end{frontmatter}



\section{Introduction} \label{sec:introduction}

The field of Human Activity Recognition (HAR) has witnessed a shift in recent years with the emerging dominance of deep learning methodologies, surpassing the capabilities of traditional techniques \cite{yang2015deep,nils2016convolutionswearables,morales2016transfer,wang2019survey}. The ability of deep learning based techniques to extract complex non-linear patterns autonomously from raw data positions it at the forefront of HAR. However, the use of deep learning brings a new and unique set of challenges. 

The first major challenge is that the success of traditional deep learning techniques is contingent on the availability of extensive labeled datasets: a requirement that presents several difficulties in the context of smartphone based HAR. Collecting a large corpus of labeled data in this domain presents several challenges, including the privacy/sensitive nature of smartphone data, the costs associated with data annotation, and the heterogeneity in devices, sensors, user behaviors, and environmental conditions \cite{duffy2019annotation,nweke2018challenges}. These data requirements may have been mitigated by recent results leveraging the sophisticated reasoning capabilities of Large Language Models (LLMs) for zero- or few-shot learning \cite{ji2024llm}; however, there is evidence that existing models have memorized major datasets, muddying their apparent success \cite{haresamudram2024llm}. More recent work has demonstrated that LLMs can be adapted for time-series tasks like HAR by aligning sensor data with textual representations and refining the model through task-aware tuning \cite{li2024llm}. Due to the large computational requirements of these models, in this paper we focus on smaller networks that can be run on portable devices.

Recent work has made progress in overcoming these challenges by learning representations from large amounts of unlabeled data in a semi-supervised way \cite{thakur2022unsupervised,saeed2019}. A notable recent example of this approach is \cite{saeed2019}, where a model was trained in a self-supervised way to identify specific transformations in an extensive corpus of unlabeled data. The base layers were then transferred to a new model and frozen, which then underwent supervised training on a comparatively smaller labeled dataset. This approach resulted in semi-supervised models that matched or exceeded the performance of previous models trained using substantially greater amounts of labeled data.

A second challenge faced by deep learning based techniques is their susceptibility to minute perturbations in input data. Such perturbations, which may be imperceptible to a human, can result in erroneous yet confident predictions. These perturbed samples are often called \textit{adversarial samples} \cite{fgsm,tygar2011adv,papernot2016rec,mcdaniel2016,jha2018distance}. Adversarial samples, being a natural consequence of the large dimensionality of the input space \cite{khoury2018,chattopadhyay2019}, may be an unavoidable aspect of many current approaches to machine learning \cite{gal2019sufficient}. The existence of adversarial samples underscores a fundamental deficit in the robustness of neural networks, impeding their deployment in scenarios where reliability or safety are crucial. This represents a significant barrier to the deployment of deep learning based techniques in many real-world applications.

To fortify models against these adversarial threats, a range of strategies have been developed. In \cite{hendrycks2019ssrobustness}, it was demonstrated that a self-supervised training stage can improve model robustness by learning more robust representations, with significant gains in accuracy on highly perturbed images. Adversarial training, a method that integrates adversarial samples into the training dataset, has proven effective in broadening the model's comprehension of the input space \cite{kurakin2017adversarial,vaishnavi2019,bai2021adv,wang2023adv}. Nonetheless, it is important to recognize that adversarial training does not eradicate adversarial samples entirely, but rather improves robustness against specific adversarial attack types employed during training within a determined perturbation boundary \cite{song2018,li2019nr,bai2021adv,yu2022strengthat}. The increased robustness is valuable, but the adversarial samples which persist outside of the region in which adversarial examples were generated during training continue to undermine the model's reliability. It is also important to take additional steps to increase model reliability.

Uncertainty Quantification (UQ) presents an alternative and complementary approach for building more robust and reliable models. UQ models take a Bayesian approach to the data and model weights and aim to generate predictions accompanied by corresponding uncertainties (which should be high for adversarial and out-of-distribution samples). Models developed with these techniques often have superior prediction accuracy to traditional techniques and are more robust against overfitting - especially in the case of limited data \cite{gal2015bayesian}. By leveraging the model's reported uncertainty, control mechanisms can be developed (at the most simple, discarding samples with uncertainties above some threshold) which significantly enhance model accuracy, robustness and reliability.

For example, in \cite{lee2018control}, the authors integrate Bayesian neural networks (BNN) with model predictive control to enhance the safety of imitation learning. The framework uses uncertainty metrics from the BNN to detect out-of-distribution inputs, prompting a switch to a more reliable control method for highly uncertain inputs. In \cite{kanazawa2022} the authors propose an uncertainty-aware reinforcement learning framework that uses an ensemble of distributional critics to estimate both epistemic and aleatoric uncertainties. Their UA-DDPG method improves exploration and risk-sensitive policy learning in continuous control tasks. Approaches such as these underscore the practical value of uncertainty estimates in safety, training and exploration. 

However, the benefits of UQ techniques do not come for free. Often UQ involves running inference multiple times and considering the results to be samples representing some underlying distribution (e.g. see model ensembles, Monte Carlo Dropout, Bayes by Backprop, Variational Inference, and others \cite{lakshminarayanan2017ensembles,gal2016mcd,graves2011vi,blundellbbb,shridhar2019acg}). These multiple samples scale the cost of inference with the number of samples. The computational demands of UQ mean that it is impractical in applications with computational restrictions: for example, on portable devices with small batteries, or low-power hardware. Unfortunately, this is an important use case for HAR models.

To circumvent these computational constraints, Ensemble Distribution Distillation (EDD) \cite{malinin2019edd} offers a compelling solution. EDD trains a single model as a Prior Network \cite{malinin2018priors} (i.e. a model whose output is interpreted as parameterising a distribution modelling the posterior) to emulate the output of a larger ensemble by minimizing the Kullback–Leibler divergence between the modeled distribution and the ensemble's outputs. In contrast to other techniques mentioned above, which require multiple stochastic forward passes to approximate the predictive distribution, EDD distills the behavior of a full ensemble into a single network that outputs a parameterized distribution, thereby enabling one-pass inference. This approach may increase the complexity, or computational requirements of training the model, but in exchange allows a single model to effectively exploit the benefits of ensemble methods, including uncertainty quantification and enhanced predictive accuracy, while incurring almost no increased computational cost at inference time.

The gains in both efficiency and robustness of this approach is particularly critical in real-world applications like fall detection systems. In such settings, rapid and reliable decision-making is essential, and the ability to adjust control measures based on precise uncertainty estimates can significantly enhance system performance and reliability. This dynamic control ensures that the system remains both accurate and resilient, even under challenging conditions.

In the present work, we explore the use of EDD to enhance the semi-supervised training of deep neural networks for HAR. We demonstrate that our approach to EDD for HAR significantly improves both the accuracy and robustness of deep learning models without increasing data requirements or inference costs. Our model is able to retain the predictive accuracy and high quality total uncertainty estimates of an ensemble even under heavy adversarial perturbation of the inputs. While our approach also provides principled separable measures of uncertainty (epistemic and aleatoric), our primary focus in this work is demonstrating how EDD enhances model accuracy and resilience to adversarial samples in HAR. Future work may explore how these separable uncertainties could further improve decision-making and control in real-world scenarios. Our claims are supported by extensive evaluations on several publicly available datasets.

Our main contributions are:
\begin{itemize}
    \item A novel method for obtaining principled uncertainty estimates without large amounts of labeled data.
    \item The development of a self-supervised EDD framework specifically tailored for HAR.
    \item An innovative data augmentation technique for EDD using time series data.
    \item Empirical evidence showing improved model accuracy, robustness and reliability in various settings.
\end{itemize}
The benefits of our approach include:
\begin{itemize}
    \item Improvements in model accuracy.
    \item Significantly increased resilience to adversarial samples.
    \item Principled uncertainty estimates that can accurately predict incorrectly labeled samples.
\end{itemize}

By addressing the challenges of limited labeled data and model uncertainty in HAR, our work offers a scalable and efficient solution that advances the state of the art and has practical implications for real-world deployments.

The paper is organized as follows: in Section \ref{sec:prelims} we give an overview of related works and methodologies that the current work builds on; in Section \ref{sec:approach} we introduce the proposed Ensemble Distribution Distillation framework, including network architecture and training implementation; in Section \ref{sec:evaluation} we describe our evaluation methodology and results; and finally, Section \ref{sec:conclusion} provides our conclusion, and suggests future research directions.

\section{Preliminaries and Prior Work} \label{sec:prelims}

In this section we provide a summary of prior work which provides the basis for the current paper. Specifically, we describe semi-supervised learning in the context of HAR, UQ with BNNs, and EDD.
\subsection{Semi-Supervised Learning} \label{subsec:semisuplearning}
Semi-supervised learning techniques, where a learning algorithm is able to make use of both labeled and unlabeled data, is an active and exciting area of research. Semi-supervised learning techniques are of particular use in domains where unlabeled data are abundant, but labeled instances are rare or expensive. Indeed, owing to the sharp increase in available data enabled by the use of unlabeled instances, semi-supervised techniques have been able to achieve performance that would not be possible with traditional fully-supervised techniques in a number of fields owing to the relative lack of labeled data, including in computer vision \cite{laina2019vision}, robotics \cite{nair2019cig}, audio processing \cite{wav2vec}, recommendation systems \cite{jiang2023ss} and, quite famously, natural language processing \cite{kojima2022llm}.

In a recent advance in the field of self-supervised learning, \cite{saeed2019} introduced a novel self-supervised learning framework designed specifically for the task of HAR which was able to significantly improve the performance of a convolutional neural network trained on a small amount of labeled accelerometer data by using features learned on a large amount of unlabeled data in a self-supervised way. 

In the approach used by Saeed et al. \cite{saeed2019}, two networks are used with identical convolutional layers as their base: one for the self-supervised task, and one for the fully-supervised activity recognition task. To generate the training data for the self-supervised task, each of 8 transforms are applied (noising, scaling, rotation, negation, time-reversal, window permutation, time-warping and channel shuffling) to the unlabeled instances, and the generated instances are labeled according to which transform has been applied (including labeling the original instance as no transform). That is, for each transform $T$ and each instance $x$, we create a pair of labeled data:
\begin{equation}
    \{(T(x),\,1),\ (x,\,0)\}.
\end{equation}
The self-supervised network consists of fully-connected layers specific to each transformation, connected to a shared convolutional base. The network is then trained with these synthetic data to predict the applied transformation, simultaneously for all transformations.

\begin{figure}[h]
    \centering
    \includegraphics[width=0.7\textwidth]{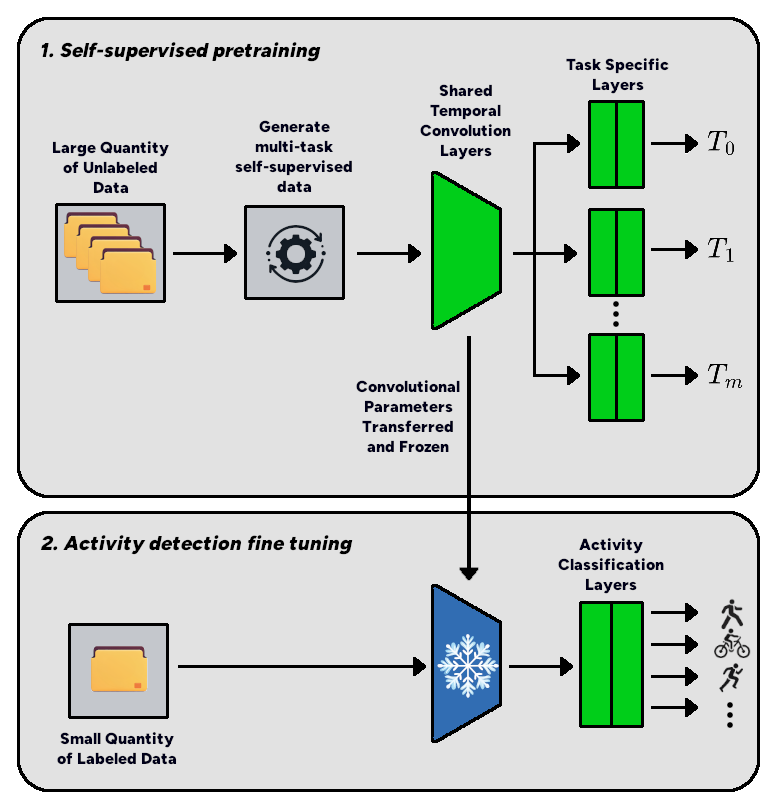}
    \caption{Schematic representation of the semi-supervised the approach, reproduced from \cite{saeed2019}. In the initial self-supervised step, a convolutional neural network is trained to predict the applied transformation. The learned features are then used in the downstream fully-supervised step to train a HAR model using a small labeled dataset.}
    \label{fig:schematic}
\end{figure}

For an activity recognition (fully-supervised) task, a smaller number of labeled instances are used as training data. Once the training has converged in the self-supervised step, the weights from the convolutional layers in the self-supervised network are transferred to the network for the activity recognition task and some (or all) of the convolutional layers are frozen. Next, this network is trained on the labeled instances in a fully-supervised way. See Figure \ref{fig:schematic} for a schematic representation.

In the present work, we extend this approach to train a network in a semi-supervised way to not only perform activity recognition, but to provide well-calibrated uncertainty information as well.

\subsection{Uncertainty Quantification}\label{subsec:uq}
As models built with deep neural networks grow in power and complexity they are increasingly finding use in domains where informed decision making is critical, where acting on false predictions can result in significant financial losses, or even loss of life. Notable examples include medical diagnosis, hiring practices, finance, autonomous vehicles, to name just a few. However, despite achieving impressive performance across many domains, deep neural networks have a tendency to make overconfident predictions and are mostly unable to provide meaningful measures of uncertainty. This makes the predictions of deep neural networks untrustworthy and is a serious barrier to deployment in many use cases.

Quantifying uncertainty plays a crucial role in assessing model reliability and robustness, particularly in real-world applications where inputs may deviate significantly from training data. Uncertainty-aware models are more resilient against adversarial inputs and noise, enhancing their practical applicability.

Sources of uncertainty that result in uncertain predictions can be broken down into two categories \cite{blum2021aleatoric}:
\begin{itemize}
    \item \textit{Aleatoric} or \textit{Data} Uncertainty: Aleatoric uncertainty refers to randomness inherent to the data arising from artifacts such as lens flares in images, background noise in audio recordings, sensor noise, or other sources of uncontrollable variability. This type of uncertainty is independent of the model and therefore is \textit{irreducible}.
    \item \textit{Epistemic}, or \textit{Knowledge} Uncertainty: Epistemic uncertainty arises as a result of a model possessing incomplete knowledge about the current input and occurs in regions of the input space which are either outside of the distribution generated by the training data, or in a region which is only sparsely covered by the training data. Therefore, Epistemic Uncertainty can be reduced with more data and is thus \textit{reducible}.
\end{itemize}

Being able to not only quantify, but to differentiate between sources of uncertainty plays a key role in understanding both the degree of model uncertainty and its underlying reasons. This can help develop a deeper understanding of the model's behavior and also guide decision making in both training and inference.

BNNs bring the possibility of UQ by incorporating uncertainty into their design. BNNs are able to not only make point estimates for predictions (just as a traditional deep learning model), but also to quantify the associated uncertainty for each prediction. This allows for better decision making under uncertain conditions, as the predicted uncertainties allow users or control mechanisms to assess the reliability of the model's predictions and make appropriately informed decisions. BNNs can also be more robust to overfitting, and better at generalizing when compared to a traditional model of a similar size trained on the same data \cite{mitros2019ontv}.

At a high level, BNNs integrate Bayesian statistics into the framework of neural networks by representing the parameters of a neural network as random variables drawn from probability distributions instead of as single point values. This allows them to capture the uncertainty associated with each parameter and thus make more robust and informed predictions. To start, some prior distribution $p(\te)$ is defined for the parameters $\te$, reflecting the prior belief of the parameter values before seeing any data. Next, a calculation of the likelihood of seeing the dataset $\mathcal{D}$ given the parameters $p(\mathcal{D}|\te)$ is found. Finally, using Bayes' theorem, the posterior can be calculated as:
\begin{equation}\label{eq:bayes}
    p(\te | \mathcal{D}) = p(\mathcal{D}|\te)\,p(\te).
\end{equation}
Since the total uncertainty is the sum of epistemic and aleatoric uncertainties, we can be use this approach to define the uncertainties as
\begin{subequations}\label{eq:uncs}
    \begin{align}
        U_{\rm total}&=I[p(y|x,\mathcal{D})],\\
        U_{\rm aleatoric}&=\mathbb{E}_{p(\te|\mathcal{D})}(I[p(y|x,\te)]),\\
        U_{\rm epistemic}&=U_{\rm total} - U_{\rm aleatoric},
    \end{align}
\end{subequations}
where $I$ is some uncertainty measure, such as variance or entropy \cite{malinin2019edd}. Throughout the present work, we use entropy as this uncertainty measure.

In general, it is intractable to compute the posterior exactly, so approximating methods must be used. Many powerful tools have been successfully deployed for this purpose, including Markov Chain Monte Carlo (MCMC) techniques \cite{speagle2020mcmc}, Variational Inference (VI) \cite{blei2017VI}, Monte Carlo Dropout \cite{gal2016mcd}, among others. In this work, we will be focusing on ensemble techniques \cite{chen2019deepensembles}. For an in-depth review of the state of UQ research, see \cite{abdar2021review}.

Ensemble methods are a conceptually simple approach to approximating an intractable posterior distribution. Ensemble methods combine the predictions of a collection of multiple models (the ensemble) to enhance the reliability and accuracy of the system in a prediction or regression task. This approach has been empirically demonstrated to be highly effective in preventing overfitting, improving generalization ability and improving performance \cite{lakshminarayanan2017ensembles,ovadia2019canyoutrust}. The success of an ensemble relies on diversity of its members. To ensure this diversity the individual models are trained independently, with differing random initialization and mini-batch sampling for each model.

Ensembles can approximate the posterior of a BNN (including separable uncertainty estimates) by considering the prediction of each of their members as a random sample of the posterior. For samples which are near to the training data the individual models should largely agree, so there will be relatively low spread in the ensemble predictions. In contrast, samples which are far from the training set (i.e. out of distribution samples) should produce a wide spread in predictions.

Owing to their effectiveness and relatively simple implementation, ensemble methods have been used extensively to estimate uncertainty in deep learning applications, \cite{lakshminarayanan2017ensembles,chen2019deepensembles,swaroop2020pointclouds}. However, ensemble methods come with significant increases in computational and memory requirements. Since each ensemble member is independent and must be run on each sample to generate the uncertainty estimates, an ensemble with $M$ members essentially scales the computational demands and memory requirements at inference time by a factor of $M$; rendering them impractical in many applications. It is possible to reduce this requirement by sharing the base layers of the network and having a smaller number of independent layers at the end \cite{valdenegrotoro2019deepsf} but while this reduces the additional resource requirements, it may still be infeasible in low power devices.

\subsection{Ensemble Distribution Distillation} \label{subsec:edd}
There has been interest in knowledge distillation of neural networks for several years now \cite{hinton2015distilling,furlanello2018born}. Using the Kullback–Leibler (KL) divergence, which is a measure of dissimilarity between two distributions, Ensemble
   Distillation is a technique which minimizes the difference the expectation of an ensemble and the output of the distilled model \cite{gou2021knowledge,anil2018large}. This approach is also sometimes called mixture distillation. This effectively trains a model to capture the predictive mean of an ensemble and has successfully been used to improve predictive performance. Additionally, the process does capture some uncertainty information - the distilled model is able to capture the ensemble's \textit{total} uncertainty. However, information about the ensemble's diversity is lost, and therefore it is no longer possible to obtain separate estimates of total aleatoric and epistemic uncertainties.

In \cite{malinin2019edd} a new approach was proposed which is capable of distilling an ensemble's mean and diversity into a single model. This has been further investigated and generalized in \cite{lindqvist2020agf}. In this approach, the outputs of the distillation model are considered to be the parameters of a probability distribution. The distillation model is trained by minimizing the KL divergence between the distribution parametrized by the model's outputs and the distribution implied by the ensemble's diversity.

More precisely stated, given a sample, $x$, and parameters, $\ph$, the outputs, $\al$, of the distillation model, $f(x;\phi)$, are considered to be the parameters of a distribution which models the distribution from which the ensemble members are sampling. That is, the ensemble output, which is a set of samples from an implicit distribution, is distilled into an explicit distribution modeled by a single neural network:
\begin{equation}
    \left\{\texttt P(y|x;\te^{(m)})\right\}_{m=1}^M\rightarrow \texttt p(\mathbf{\pi}|x;\ph).
\end{equation}

In this way, by choosing an appropriate prior for the parametrized distribution, the distillation model is able to emulate an ensemble without the additional resource requirements at inference time.

\begin{figure}[h]
    \centering
    \begin{tikzpicture}[
        font=\sffamily,
        line cap=round,
        line join=round,
        >=Latex,
        arrow/.style={->, thick},
        scale=0.5
    ]
    
        \node[
            rounded corners,
            minimum width=2cm,
            minimum height=1cm,
            align=center,
            font=\scriptsize
        ] (ensemble) 
        {
            \begin{tikzpicture}[scale=0.4]
                \node at (-0.00, 1.00)        {\includegraphics[width=0.75cm]{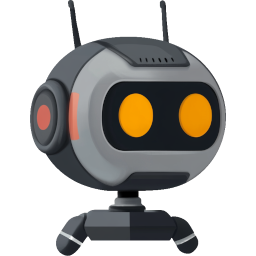}};
                \node at ( 0.70, 0.50)        {\includegraphics[width=0.75cm]{bot.png}};
                \node at ( 1.50, 0.00)        {\includegraphics[width=0.75cm]{bot.png}};
    
                \node at (-0.75, 0.50)        {\includegraphics[width=0.75cm]{bot.png}};
                \node at ( 0.00, 0.00)        {\includegraphics[width=0.75cm]{bot.png}};
                \node at ( 0.75,-0.50)        {\includegraphics[width=0.75cm]{bot.png}};
    
                \node at (-1.50, 0.00)        {\includegraphics[width=0.75cm]{bot.png}};
                \node at (-0.75,-0.50)        {\includegraphics[width=0.75cm]{bot.png}};
                \node at ( 0.00,-1.00)        {\includegraphics[width=0.75cm]{bot.png}};
            \end{tikzpicture}\\[2pt]
            Pre-trained Ensemble
        };
    
        \node[
            rounded corners,
            minimum width=3.5cm,
            minimum height=1.75cm,
            align=center,
            below=1.5cm of ensemble,
            font=\scriptsize
        ] (distil) 
        {
            \begin{tikzpicture}[scale=0.6]
                \node at (0,0)        {\includegraphics[width=0.75cm]{bot.png}};
            \end{tikzpicture}\\[-2pt]
            Distillation Model
        };
    
    
        \node[right=3.5cm of ensemble] (topDistAnchor) {};
        \begin{scope}[shift={(topDistAnchor)}]
            \pgfmathsetseed{42};
            \foreach \x in {-1.5, -1.4, -1.3, -1.2, -1.1, -1.0, -0.9, -0.8, -0.7, -0.6, -0.5, -0.4, -0.3, -0.2, -0.1, 0, 0.1, 0.2, 0.3, 0.4, 0.5, 0.6, 0.7, 0.8, 0.9, 1.0, 1.1, 1.2, 1.3, 1.4, 1.5} {
                \pgfmathsetmacro{\randy}{0.15*(rand - 0.5)}
                \draw[fill=red, draw=red] (\x,{exp(-(1.5*\x)^2) - 0.2 + \randy}) circle (1.3pt);
            }
            \node[above=0.25cm, align=center, font=\scriptsize] at (0,0.8) {Ensemble members each sample from\\an underlying distribution};
        \end{scope}
    
        \node[right=3.4cm of distil] (bottomDistAnchor) {};
    
        \begin{scope}[shift={(bottomDistAnchor)}]
            \draw[thick] plot [smooth, tension=1] coordinates {(-1.6,-0.5) (-1.2,0.1) (-0.8,-0.3) (-0.4,0.5) (0,-0.1) (0.4,0.7) (0.8,-0.3) (1.2,0.1) (1.6,-0.5)};
            \node[below=1cm, align=center, font=\scriptsize] at (0,0.8) {Distillation model predicts a\\distribution instead of point estimates};
        \end{scope}
    
        \node[shift={(3.5cm,-1.85cm)}, right=of topDistAnchor] (finalDistAnchor) {};
        \begin{scope}[shift={(finalDistAnchor)}]
            \draw[thick] plot [smooth, tension=1] coordinates {(-1.6,-0.20)  (-1,-0.09)  (-0.5,0.37)  (0,0.8)  (0.5,0.37)  (1,-0.09)  (1.6,-0.20)};
            \pgfmathsetseed{42};
            \foreach \x in {-1.5, -1.4, -1.3, -1.2, -1.1, -1.0, -0.9, -0.8, -0.7, -0.6, -0.5, -0.4, -0.3, -0.2, -0.1, 0, 0.1, 0.2, 0.3, 0.4, 0.5, 0.6, 0.7, 0.8, 0.9, 1.0, 1.1, 1.2, 1.3, 1.4, 1.5} {
                \pgfmathsetmacro{\randy}{0.15*(rand - 0.5)}
                \draw[fill=red, draw=red] (\x,{exp(-(1.5*\x)^2) - 0.2 + \randy}) circle (1.3pt);
            }
            \node[below=1cm, align=center, font=\scriptsize] at (0,0.8) {Distillation model learns\\distribution which reproduces\\ensemble diversity and\\uncertainty measures};
        \end{scope}
    
    
        \draw[arrow] (ensemble.east) -- ++(0.3,0) |- ($(topDistAnchor.west) + (-1.75,0)$);
    
        \draw[arrow] (distil.east) -- ++(0.3,0) |- ($(bottomDistAnchor.west) + (-1.75,0)$);
    
        \draw[<-, thick] ($(bottomDistAnchor)+(0.0,0.5)$) -- node[midway, left=2pt, align=center, font=\scriptsize] {Train to minimize\\KL divergence} ($(topDistAnchor)+(0.0,-0.5)$);
    
        \draw[arrow] ($(topDistAnchor)+(1.8,0)$) -- ($(finalDistAnchor)+(-1.85,0.25)$);
        \draw[arrow] ($(bottomDistAnchor)+(1.8,0)$) -- ($(finalDistAnchor)+(-1.85,-0.25)$);
    \end{tikzpicture}
    
    \caption{The Ensemble Distribution Distillation procedure. First, an ensemble of models is trained, and each member effectively samples from the true underlying distribution. The distillation model, whose outputs parametrize a family of distributions, is trained to minimize the KL divergence between its own prediction and the distribution implied by the ensemble. Once training has converged, the distillation model produces a distribution which closely matches the distribution implied by the ensemble without requiring multiple passes.}
    \label{fig:EDD}
\end{figure}

A summary of the technique is as follows (Figure 2). First, an ensemble of \( M \) models is developed and trained in the usual way. Next, the trained ensemble is used to generate the \textit{distillation dataset}, \( \mathcal{D}_{\rm ens}=\{x^{(i)}, \mathbf{\pi}^{(i,1:M)}\}_{i=1}^{N} \), which consists of input samples \( x^{(i)} \) along with the corresponding outputs, \( \mathbf{\pi}^{(i,1:M)} \), produced by the ensemble. Finally, given this dataset, the distillation model is trained to match the ensemble’s predictive distribution by minimizing the negative log-likelihood (NLL) of the ensemble outputs:
\begin{equation}
    \mathcal{L}(\phi,\mathcal{D}_{\rm ens})=-\mathbb{E}_{\hat{p}(x)}\left[\mathbb{E}_{\hat{p}(\mathbf{\pi}|x)}\ln p(\mathbf{\pi}|x;\phi)\right].
\end{equation}
We minimize the NLL instead of the KL divergence since they differ only by a term which is constant with respect to model parameters; therefore minimizing either one is equivalent to minimizing the other. The resulting model is then able to reproduce both the predictions and the separable uncertainty measures of the ensemble.

In this work, the family of distributions parametrized by the prior network is a Dirichlet distribution over the simplex of activity classes. Given this distribution, we are able to calculate separable measures of uncertainty from the single model by using equations \eqref{eq:uncs} with entropy as the uncertainty measure. Specifically, given the distillation model's output $\al$ over $K$ classes, we define
\[S=\sum_{i=1}^K\al_i,\,\, \mu_i=\frac{\al_i}{S}, \]
so that
\begin{subequations}
    \begin{align}
        U_{\rm total}&=-\sum_{i=1}^K\mu_i\,\log(\mu_i),\\
        U_{\rm aleatoric}&=-\sum_{i=1}^K\mu_i\left(\psi(\al_i)-\psi(S)\right),\\
        U_{\rm epistemic}&=U_{\rm total} - U_{\rm aleatoric},
    \end{align}
\end{subequations}
where $\psi$ is the \textit{digamma function}.

\section{Semi-Supervised Ensemble Distillation Approach} \label{sec:approach}
Our method distills ensemble diversity into a single robust prior network using EDD, significantly improving classification accuracy and adversarial robustness. As an additional capability, this method also provides principled separable uncertainty estimates (epistemic and aleatoric); although exploring the full practical utility of these estimates is not within the primary scope of this study.

We train an ensemble on a small amount of labeled data and a relatively larger amount of unlabeled data by following the approach in \cite{saeed2019} to train each ensemble member. Each member of the ensemble is trained independently, using the same limited set of labeled samples, though in different batches. We then use data augmentation techniques to produce a new dataset from the entire labeled and unlabeled training sets consisting of augmented samples as data and the ensemble outputs on that data as labels. We then use EDD to train the prior network to predict the distribution underlying the ensemble's outputs following the approach outlined in \cite{malinin2019edd}. Our approach is able to not only successfully capture the superior predictive accuracy of the ensemble, but also the uncertainty information -- achieving vastly improved accuracy as well as benefits to robustness, without any change to the model's architecture or increase in inference costs.

For the sake of clear comparison, the training procedure and architectures of each member of the ensemble, as well as the architecture of the prior network itself, match the architecture described in \cite{saeed2019}. This demonstrates that the benefits to performance and robustness are due only to the leveraging of the ensemble's diversity.

\subsection{Training}\label{subsec:training}
In this subsection, we will individually introduce the ingredients used in the training process. We show how they are pieced together in Algorithm \ref{algo:training}.

\subsubsection{Data Augmentation}
In \cite{malinin2019edd} it was demonstrated that while EDD was effective at correctly decomposing total uncertainty into epistemic and aleatoric uncertainty it failed to capture epistemic uncertainty in regions away from the training data. For this reason, additional inputs were sampled away from the training data and added to the training set. To generate auxiliary training data in the HAR setting, we use two different approaches to data generation.

Since we want our distilled model to be robust against transformations under which the data should be invariant, as well as capture epistemic uncertainty otherwise, we experiment with augmenting the training data with the same transforms used in the self-supervised step. For each instance, $x$, in the unlabeled data and for every transform, $T$, we augment the distillation dataset with the transformed instance, $T(x)$.

Additionally, we expect the model should return low uncertainty on data near to the training set, high uncertainty on random noise and return an increasing uncertainty in the region in-between. Furthermore, since the datasets in our experiments consist mostly of participants intentionally performing tasks which fall unambiguously into one of the classes, a distillation model trained on the data may not correctly capture the uncertainty of samples which fall \textit{between} classes. In order to address these issues we augment the training set with \textit{weighted combinations} of samples from the training data. That is, for some weight parameter, $r$, and $N$, randomly selected samples, $\{x_i\}$, we add to the auxiliary training data the new sample, $y$, where 
\begin{equation}\label{eq:combos}
    y = \frac{\sum_{i=0}^{N-1}r^i\,x_i}{\sum_{i=0}^{N-1}r^i}.
\end{equation}
Equation \eqref{eq:combos} was designed to generate new samples which act as intermediate states between one or more activity classes. By appropriately choosing $r$ and creating samples for a range of $N$, the auxiliary dataset is extended into regions which share features with two or more of the activity classes. This gives the model greater exposure to samples on which the ensemble returns high degrees of uncertainty.

\subsubsection{Annealing}
Temperature annealing plays a crucial role in EDD as it helps to ensure the stability of training in the early stages \cite{saeed2019}. During training, the ensemble and distribution model are given a temperature parameter, $T$, and an annealing schedule is used to gradually decrease $T$ for both the model and the target ensemble during training. This is to increase common support during early stages and reemphasize diversity once the model has started to converge to the desired distribution.

Similarly, we found that including a large number of samples, generated as weighted combinations in the early stages, also negatively impacts training. Including a large number of uncertain samples encourages the model in the early stages to bias towards the center of the simplex (i.e. high uncertainty). We hypothesize that this reduces common support during later stages once the temperature is lower and causes instability in training. For this reason, at each epoch we gradually increase the maximum number of samples that can be used to create the augmented data, starting at 1 (i.e. no augmentation).

\subsubsection{Pre-training}
In \cite{hendrycks2019ssrobustness} it was shown that using a self-supervised step can itself improve both robustness and uncertainty. For this reason, we experiment with including the pre-trained weights for the convolutional layers from the pretext task in our EDD model. Our experiments included the use of no pre-trained parameters, as well as using pre-trained parameters with 0, 1, or 2 frozen layers.

\subsubsection{Training Procedure}
The procedure for training our distillation model is as follows: 

First, following the architecture and procedure of \cite{saeed2019}, train 50 semi-supervised models using an unlabeled dataset, $D_U$, on the pretext task and a smaller labeled subset, $D_L$, for the downstream task. These models are then collectively considered the ensemble we wish to distill using EDD.

Next, using the transforms, $T$, used in the pretext task, create an augmented dataset, $D_A$, consisting of all instances from $D_U$, as well as the result of each transform. Finally, we train the distillation model to minimize the KL divergence between itself and the ensemble outputs, further augmenting the data as the epochs proceed by increasing the number of weighted combinations of samples appearing in the training data. This is summarized in Algorithm \ref{algo:training}.

In our experiments, we found that the best results came from using both transformations and weighted combinations, and pre-trained weights without frozen layers. This is the procedure we use for the results in the following section.

\begin{algorithm}[H]
\footnotesize
\caption{Training the prior network using EDD with pretraining and data augmentation}
\label{algo:training}
\SetKwInOut{Input}{Input}
\SetKwInOut{Output}{Output}

\Input{Unlabeled instance set $D_U$, number of epochs $E$, transforms $T$, annealing rate $v_t$, initial temperature $t_0$, maximum temperature $t_{\rm max}$, maximum number of combos $N_C$, rate to increase combos in data $v_C$, combo weight parameter $r$, pre-trained weights, $\hat{\theta}$, layers to freeze, $N_F$.}
\Output{Augmented dataset $D_A$, network $M$ with parameters $\theta_M$ trained with Ensemble Distribution Distillation to capture the behavior of an ensemble trained in a semi-supervised way on limited data.}

\tcp{Add transformed data to the dataset}
\For{each instance $x$ in $D_U$}{
    Insert $x$ into $D_A$.\\
    \For{each transform $\tau$ in $T$}{
        Insert $\tau(x)$ into $D_A$.
    }
}
\If{using pre-trained weights}{
    Initialise network $M$ with pre-trained weights $\hat{\theta}$.\\
    Freeze the first $N_F$ convolutional layers.
}
\tcp{Train the distillation model}
\For{each epoch $e$ from 1 to $E$}{
    Set temperature $t$ to $t_0 - v_t\,e$, clamping to between $1$ and $t_{\rm max}$.\\
    Set maximum combo depth $C$ to be $\floor{v_c\,e}$.\\
    Randomly sample a batch $B$ of $N$ samples from $D_A$.\\
    \tcp{Add combos to the mini-batch}
    \For{each sample $x$ in $B$}{
        \tcp{Generate a weighted combination sample as in Equation \ref{eq:combos}}
        Set $W$ to 1.\\
        Choose a random number $n$ from $0$ to $C$.\\
        \For{each $i$ from $1$ to $n$}{
            Sample a random $y$ from $D_A$.\\
            Set the weight $w$ to $r^i$.\\
            Set $x$ to $x + w\,y$.\\
            Set $W$ to $W + w$.
        }
        Set $x$ to $\frac{x}{W}$.
    }
    Compute the model's predictions $y$ on $B$, using temperature $t$.\\
    Compute the ensemble outputs $\hat{y}$ on $B$.\\
    Update $\theta_M$ by descending along the gradient of the loss function $\mathcal{L}(y, \hat{y})$.
}
\end{algorithm}

\subsection{Data Preparation}\label{subsec:datapreparation}
To assess the robustness of our models, we evaluated the performance of our models against adversarial samples generated using the Fast Gradient Sign Method (FGSM) \cite{fgsm}. FGSM was selected for its simplicity, efficiency, and ubiquity. Its one-step, gradient-based approach allows for rapid generation of adversarial examples, and its broad  familiarity among the community makes it a practical baseline for benchmarking model robustness.

To enable meaningful comparisons of the impact of perturbations across different datasets, we rescale each dataset such that the mean standard deviation across all channels for all training instances is standardized to 1.0. We do this separately for accelerometer and gyroscope data.

\section{Evaluation of Semi-Supervised Ensemble Distillation} \label{sec:evaluation}
In this section, we evaluate our approach using a collection of publicly available datasets for HAR. Our experiments demonstrate gains in accuracy, improved quality of uncertainty estimates and robustness to adversarial samples. The use of these widely adopted datasets enables others to clearly see how our method performs in comparison to other standard approaches.

\subsection{Implementation} \label{subsec:architecture}
In our experiments we used 50 instances per class as the labeled training data. We used ensembles consisting of 10 and 50 members, with each member being trained independently following the training procedure and the architecture outlined in \cite{saeed2019}. The only alteration being that for the sake of variety each member had its number of hidden units scaled by a random variable, $r$, with $r$ uniformly distributed between $0.75$ and $1.25$.

The distillation network in our experiments followed the same architecture. However, instead of applying a softmax to the final logits, $x$, to represent the likelihood of the sample belonging to each class, we calculated $\alpha = \exp(x/T)$ using a temperature , $T$. We then interpret $\alpha$ as the parameters of a Dirichlet distribution over the simplex of classes. This model is trained to minimize the KL divergence between this Dirichlet distribution and the 50-member ensemble's predictions on the distillation dataset. The resulting distribution provides the model's expectation, as well as its uncertainty.

\subsection{Datasets} \label{subsec:datasets}
In order to demonstrate the resilience of our technique across a range of device types, data collection methods, and HAR tasks in diverse environments, we explore four publicly available datasets. Below, we provide a brief overview of each dataset, emphasizing their main features.

\subsubsection*{HHAR Dataset}
The Heterogeneity Human Activity Recognition (HHAR) dataset \cite{stilson2015hhar} was developed to investigate the effects of sensor and device heterogeneities in HAR. The dataset contains data from 36 devices, spanning 13 models, from 4 manufacturers. The devices recorded data at significantly different sampling rates: from 50Hz to 200Hz. The participants each carried eight smartphones in a tight pouch around the waist and wore two smartwatches on each arm. The dataset consists of sensor readings from accelerometers and gyroscopes while 9 participants performed 6 different activities: biking, sitting, standing, walking, ascending stairs, and descending stairs. Each participant performed each activity for 5 minutes to ensure equal representation.

In our experiments we use participants 1-6 as the training set, and participants 7-9 as the validation set. In order to manage the large size of this dataset, we used 1000 randomly selected samples from each user, representing an average of 1000 samples per activity in the training set.

\subsubsection*{UCI HAR}
The UCI HAR dataset \cite{garciagonzalez2020uci} contains data from 30 participants wearing a waist-mounted Samsung Galaxy S2 smartphone. The data consists of accelerometer and gyroscope signals while participants perform 6 activities: standing, sitting, laying down, walking, ascending stairs, descending stairs. 

\subsubsection*{MotionSense}
The MotionSense dataset \cite{malekzadeh2018ms} was developed to predict personal attributes in time-series sensor data generated by accelerometer and gyroscope sensors (attitude, gravity, user acceleration, and rotation rate). The data was collected at 50Hz and was sampled from an iPhone 6s kept in the participant's front pocket. There were a total of 24 participants across various ages, weights, heights and genders, performing 6 activities in the same environment and conditions: downstairs, upstairs, walking, jogging, sitting and standing. In our experiments we use participants 1-16 as the training set, and participants 17-24 as a validation set.

\subsubsection*{PAMAP2}
The PAMAP2 dataset \cite{pamap2_physical_activity_monitoring_231} was developed for activity recognition using wearable motion and heart rate sensors. The dataset consists of recordings from 9 participants performing 12 different physical activities while wearing 3 inertial measurement units (IMUs). The IMUs were positioned on the wrist, chest and ankle, capturing accelerometer, gyroscope and magnetometer data at a sampling rate of 100Hz. The dataset includes activities such as walking, running, cycling, sitting, standing and household tasks. In our experiments we use the accelerometer and gyroscope data from the IMU positioned on the wrist. We used participants 1-6 as the training set and participants 7-9 as a validation set. 

\subsection{Experiments and Results} \label{subsec:experiments}
To evaluate the capabilities of our method as compared to the previous technique \cite{saeed2019} and ensembles of the same, in each of our experiments we present results from the baseline single model, a pair of ensembles of size 10 and 50, and our proposed method. Our results demonstrate that our model offers improved prediction accuracy, improved resilience to adversarial samples and principled measures of uncertainty that remain reliable even under significant adversarial perturbation; all with no change to inference cost over the baseline single model case. All results presented in this paper are the average across 10 runs.

All our experiments were conducted on a single NVIDIA GeForce RTX 3050.

\subsubsection{Simple Accuracy}
To demonstrate the classification accuracy of our method, we present the percentage of correctly classified samples from the entire validation set for each dataset, without considering uncertainty. 

In \cite{malinin2019edd}, it was found that distilling the diversity of the ensemble into a smaller network came at the cost of a slight reduction in accuracy. Their interpretation was that some of the network's capacity had to be repurposed for the task of estimating uncertainty, so there was less capacity for discriminatory power. 

In our results however, our distilled model is able to keep pace with the ensembles in terms of accuracy on the unperturbed validation set, exceeding the single model in almost all cases. The sole exception is in PAMAP2, where the difference in accuracy between all models is negligible -- less than the standard deviation of the results. Our experiments on HHAR show a clear advantage in accuracy, with our model exceeding the accuracy of even the 10-member ensemble, and substantially exceeding the accuracy of the single model, despite their shared architecture and available training data.

Our approach keeps pace with the mean accuracy across our experiments on unperturbed data also falling within this range (Table 1).
\begin{table}[H]
    \centering
    \def\arraystretch{1.25}
    \begin{tabular}{| c | c | c | c | c |}
        \hline
        \textbf{Model} & \textbf{HHAR} & \textbf{UCI} & \textbf{MotionSense} & \textbf{PAMAP2} \\
        \hline
        \hline
        \textbf{Single Model} & $67.84\pm 1.64$ & $85.01\pm1.56$ & $83.60\pm1.97$ & $59.18\pm3.65$\\
        \textbf{Ensemble 10} & $79.74\pm 3.73$ & $89.96\pm1.57$ & $88.72\pm2.21$ & $57.92\pm2.97$\\
        \textbf{Ensemble 50} & $82.06\pm 2.04$ & $89.11\pm1.71$ & $87.73\pm1.60$ & $57.54\pm3.87$\\
        \textbf{Ours} & $82.61\pm 2.51$ & $87.06\pm0.89$ & $86.40\pm1.74$ & $58.38\pm3.53$\\
        \hline
    \end{tabular}
    \vspace{0.5em}
    \caption{Classification accuracy on unperturbed validation set, $\pm1\sigma$.}
\end{table}
These results demonstrate that our approach does not sacrifice any accuracy on the validation set and in some cases can see significant gains in line with deploying an ensemble, despite having no increase in computational requirements at inference time.

\subsubsection{Resilience to Adversarial Perturbation}
To demonstrate our method's resilience to adversarial perturbation, we present the percentage of correctly classified samples from the entire validation set with strong adversarial perturbation ($\epsilon=0.1$) for each dataset (Table 2). This is a significant perturbation, chosen to cause significant disruption to the model's ability to perform well.

Under such heavy adversarial perturbation, the single model suffers significantly -- correctly classifying samples at most around half as often as in the unperturbed case and achieving accuracies only marginally better than chance. The ensembles also see reductions in classification accuracy but fare much better, retaining the ability to correctly classify most samples despite the magnitude of perturbation.

Our distilled model sees moderate reductions in accuracy in line with the ensembles, with the average accuracy across experiments falling generally close to the accuracy of the 10 member ensemble. This demonstrates that our approach yields a substantial increase in robustness and reliability over previous techniques.

\begin{table}[H]
    \centering
    \def\arraystretch{1.25}
    \begin{tabular}{| c | c | c | c | c |}
        \hline
        \textbf{Model} & \textbf{HHAR} & \textbf{UCI} & \textbf{MotionSense} & \textbf{PAMAP2} \\
        \hline
        \hline
        \textbf{Single Model} & $37.31\pm2.48$ & $35.81\pm1.47$ & $26.44\pm3.80$ & $27.17\pm2.42$\\
        \textbf{Ensemble 10} & $62.25\pm3.65$ & $62.62\pm1.37$ & $58.29\pm3.58$ & $46.48\pm2.56$\\
        \textbf{Ensemble 50} & $70.97\pm1.62$ & $62.36\pm1.84$ & $58.84\pm3.79$ & $49.86\pm3.05$\\
        \textbf{Ours} & $63.05\pm3.83$ & $61.05\pm1.08$ & $59.28\pm3.51$ & $45.07\pm2.51$\\
        \hline
    \end{tabular}
    \vspace{0.5em}
    \caption{Classification accuracy on adversarially perturbed ($\epsilon=0.1$) validation set, $\pm1\sigma$.}
\end{table}
It is important to stress that none of these models were trained on adversarial samples and this resilience comes solely as a result of the ensemble diversity. These results demonstrate that the increased robustness arising from the ensemble diversity has been successfully distilled into our model, significantly enhancing our model's robustness and reliability.

\subsubsection{Quality of Total Uncertainty Estimates}
An effective uncertainty measure should clearly distinguish between instances where the model can confidently make accurate predictions and those where it is likely to be incorrect. This relationship will be seen in our experiments, where higher uncertainty should correlate with a greater likelihood of being misclassified. 

To demonstrate the robustness of the uncertainty estimates of our approach, we report classification accuracy for subsets comprising the bottom 25\%, 50\%, 75\%, and 100\% of samples, sorted by increasing total uncertainty, and compare the results for unperturbed and highly perturbed datasets. We also provide the multi-class AUC-ROC scores for each experiment. The AUC-ROC quantifies the effectiveness of an uncertainty measure in distinguishing correct from incorrect predictions; with a score of 1.0 indicating a perfect discriminator, 0.5 representing random chance, and 0.0 signifying complete misclassifications.

Our results demonstrate that in addition to robustness of classification accuracy, our model also achieves significantly more robust total uncertainty estimates compared to techniques described by Saeed et. al. (2019).

\subsubsection*{HHAR}
In the unperturbed case, our experiments on the HHAR dataset show that our approach yields substantial increases in accuracy at all uncertainty levels, even under significant adversarial perturbation, with performance equivalent to a 10-member ensemble (Figure 3).

While the single model sees significant reductions in both its accuracy and the reliability of its total uncertainty estimates under adversarial perturbation, the diversity of the ensembles bring significant robustness to their predictions and uncertainty estimates. Our distillation model is able to capture the benefit of this diversity.
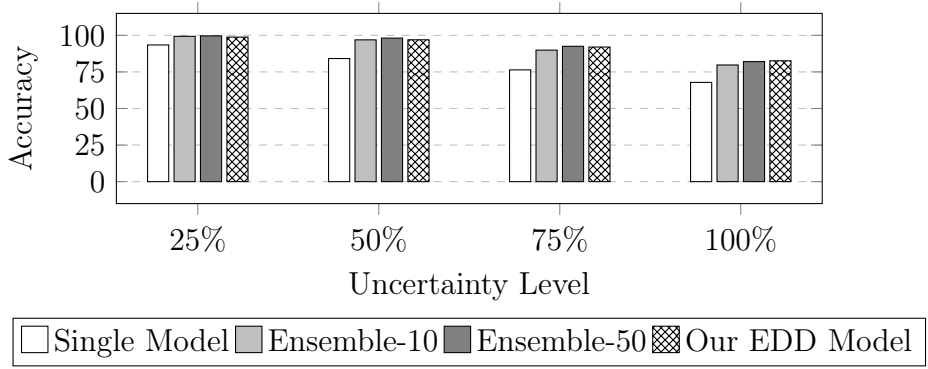
\begin{figure}[H]
    \centering
    \begin{tikzpicture}
    \begin{axis}[
        ybar,
        area legend,
        bar width=8pt,
        width=0.8\textwidth,
        height=0.3\textwidth,
        enlargelimits=0.15,
        legend style={
            at={(0.5,-0.6)},
            anchor=north,
            legend columns=-1
        },
        symbolic x coords={25\%,50\%,75\%,100\%},
        xtick=data,
        ylabel={Accuracy},
        ytick={0,25,50,75,100},
        xlabel={Uncertainty Level},
        ymin=0.0, ymax=100.0, 
        ymajorgrids=true,
        xmajorgrids=false,
        grid style=dashed,
        title={},legend image code/.code={
    \draw[#1,draw=black] (-5pt,-4pt) rectangle (5pt,6pt);
  },
    ]

    \addplot+[draw=black,fill=white]
        coordinates {(25\%,93.43) (50\%,84.15) (75\%,76.40) (100\%,67.84)};

    \addplot+[draw=black,fill=lightgray] 
        coordinates {(25\%,99.34) (50\%,96.92) (75\%,89.90) (100\%,79.74)};

    \addplot+[draw=black,fill=gray] 
        coordinates {(25\%,99.64) (50\%,98.11) (75\%,92.53) (100\%,82.06)};

    \addplot+[pattern=crosshatch,draw=black] 
        coordinates {(25\%,98.79) (50\%,96.97) (75\%,92.00) (100\%,82.61)};

    \legend{Single Model, Ensemble-10, Ensemble-50, Our EDD Model}

    \end{axis}
    \end{tikzpicture}
    \caption{Comparison of model accuracies on HHAR with no adversarial perturbation at different uncertainty levels}
    \label{fig:model_accuracy}
\end{figure}
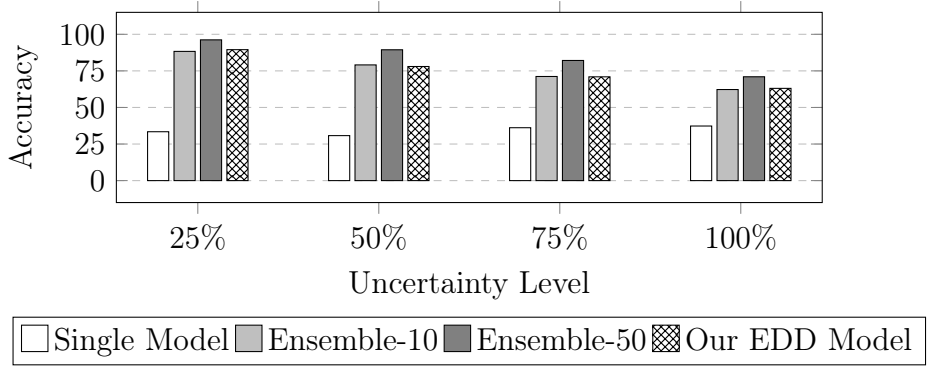
\begin{figure}[H]
    \centering
    \begin{tikzpicture}
    \begin{axis}[
        ybar,
        area legend,
        bar width=8pt,
        width=0.8\textwidth,
        height=0.3\textwidth,
        enlargelimits=0.15,
        legend style={
            at={(0.5,-0.6)},
            anchor=north,
            legend columns=-1
        },
        symbolic x coords={25\%,50\%,75\%,100\%},
        xtick=data,
        ylabel={Accuracy},
        ytick={0,25,50,75,100},
        xlabel={Uncertainty Level},
        ymin=0.0, ymax=100.0, 
        ymajorgrids=true,
        xmajorgrids=false,
        grid style=dashed,
        title={},legend image code/.code={
    \draw[#1,draw=black] (-5pt,-4pt) rectangle (5pt,6pt);
  },
    ]

    \addplot+[draw=black,fill=white]
        coordinates {(25\%,33.41) (50\%,30.75) (75\%,36.15) (100\%,37.31)};

    \addplot+[draw=black,fill=lightgray] 
        coordinates {(25\%,88.32) (50\%,79.08) (75\%,71.17) (100\%,62.25)};

    \addplot+[draw=black,fill=gray] 
        coordinates {(25\%,96.20) (50\%,89.42) (75\%,82.12) (100\%,70.97)};

    \addplot+[pattern=crosshatch,draw=black] 
        coordinates {(25\%,89.50) (50\%,77.99) (75\%,70.93) (100\%,63.05)};

    \legend{Single Model, Ensemble-10, Ensemble-50, Our EDD Model}

    \end{axis}
    \end{tikzpicture}
    \caption{Comparison of model accuracies on HHAR with heavy adversarial perturbation ($\epsilon=0.1$) at different uncertainty levels}
    \label{fig:model_accuracy}
\end{figure}

\begin{table}[H]
    \centering
    \def\arraystretch{1.25}
    \begin{tabular}{| c | c | c |}
        \hline
        \textbf{Model} & \textbf{Unperturbed data} & \textbf{Perturbed data $(\epsilon=0.1)$} \\
        \hline
        \hline
        \textbf{Single Model} & 0.7578 & 0.4443 \\
        \textbf{Ensemble 10} & 0.8567 & 0.7484 \\
        \textbf{Ensemble 50} & 0.8791 & 0.8096 \\
        \textbf{Ours} & 0.8540 & 0.7321 \\
        \hline
    \end{tabular}
    \vspace{0.5em}
    \caption{Comparison of AUC-ROC score of different models on unperturbed and highly perturbed HHAR dataset.}
\end{table}
Our distillation model also demonstrates an AUC-ROC score in line with a 10-member ensemble in both perturbed and unperturbed conditions, highlighting its capacity to effectively discriminate between correct and incorrect samples.
\subsubsection*{UCI}
In our experiments on UCI, the benefits of using ensembles on unperturbed data was statistically significant, but small. This is a reflection of the fact that the previous technique already achieved impressive results on this dataset. Our distillation model keeps pace with small but significant gains in accuracy at all uncertainty levels.

In the case of perturbed data the single model sees significant drops in accuracy at all levels of uncertainty. Our distillation model keeps pace with the ensembles on the entire perturbed dataset, as well as in the bottom 25\%, but lags behind (still ahead of the single model) when there is an intermediate uncertainty cut-off.
\begin{figure}[H]
    \centering
    \begin{tikzpicture}
    \begin{axis}[
        ybar,
        area legend,
        bar width=8pt,
        width=0.8\textwidth,
        height=0.3\textwidth,
        enlargelimits=0.15,
        legend style={
            at={(0.5,-0.6)},
            anchor=north,
            legend columns=-1
        },
        symbolic x coords={25\%,50\%,75\%,100\%},
        xtick=data,
        ylabel={Accuracy},
        ytick={0,25,50,75,100},
        xlabel={Uncertainty Level},
        ymin=0.0, ymax=100.0, 
        ymajorgrids=true,
        xmajorgrids=false,
        grid style=dashed,
        title={},legend image code/.code={
    \draw[#1,draw=black] (-5pt,-4pt) rectangle (5pt,6pt);
  },
    ]

    \addplot+[draw=black,fill=white]
        coordinates {(25\%,98.94) (50\%,97.37) (75\%,92.03) (100\%,85.01)};

    \addplot+[draw=black,fill=lightgray] 
        coordinates {(25\%,100.00) (50\%,99.61) (75\%,96.93) (100\%,89.96)};

    \addplot+[draw=black,fill=gray] 
        coordinates {(25\%,100.00) (50\%,99.53) (75\%,96.94) (100\%,89.11)};

    \addplot+[pattern=crosshatch,draw=black] 
        coordinates {(25\%,100.00) (50\%,99.74) (75\%,93.41) (100\%,87.06)};

    \legend{Single Model, Ensemble-10, Ensemble-50, Our EDD Model}

    \end{axis}
    \end{tikzpicture}
    \caption{Comparison of model accuracies on UCI with no adversarial perturbation at different uncertainty levels}
    \label{fig:model_accuracy}
\end{figure}
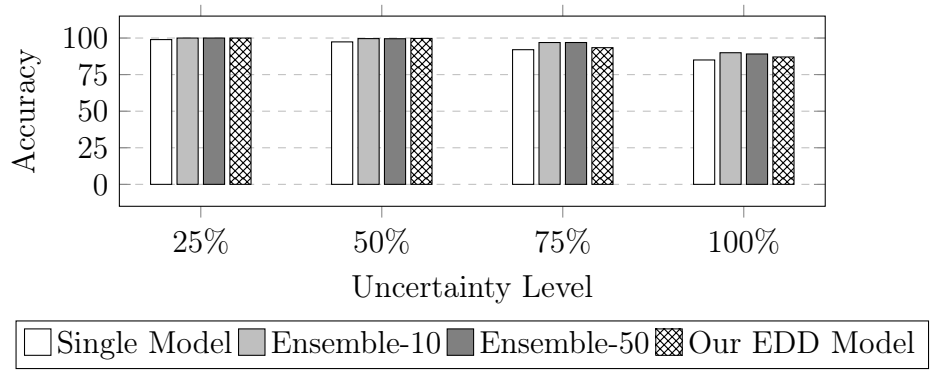
\begin{figure}[H]
    \centering
    \begin{tikzpicture}
    \begin{axis}[
        ybar,
        area legend,
        bar width=8pt,
        width=0.8\textwidth,
        height=0.3\textwidth,
        enlargelimits=0.15,
        legend style={
            at={(0.5,-0.6)},
            anchor=north,
            legend columns=-1
        },
        symbolic x coords={25\%,50\%,75\%,100\%},
        xtick=data,
        ylabel={Accuracy},
        ytick={0,25,50,75,100},
        xlabel={Uncertainty Level},
        ymin=0.0, ymax=100.0, 
        ymajorgrids=true,
        xmajorgrids=false,
        grid style=dashed,
        title={},legend image code/.code={
    \draw[#1,draw=black] (-5pt,-4pt) rectangle (5pt,6pt);
  },
    ]

    \addplot+[draw=black,fill=white]
        coordinates {(25\%,52.25) (50\%,47.12) (75\%,42.15) (100\%,35.81)};

    \addplot+[draw=black,fill=lightgray] 
        coordinates {(25\%,99.28) (50\%,93.43) (75\%,72.46) (100\%,62.62)};

    \addplot+[draw=black,fill=gray] 
        coordinates {(25\%,99.80) (50\%,95.81) (75\%,72.87) (100\%,62.36)};

    \addplot+[pattern=crosshatch,draw=black] 
        coordinates {(25\%,99.53) (50\%,80.87) (75\%,62.65) (100\%,61.05)};

    \legend{Single Model, Ensemble-10, Ensemble-50, Our EDD Model}

    \end{axis}
    \end{tikzpicture}
    \caption{Comparison of model accuracies on UCI with heavy adversarial perturbation ($\epsilon=0.1$) at different uncertainty levels}
    \label{fig:model_accuracy}
\end{figure}
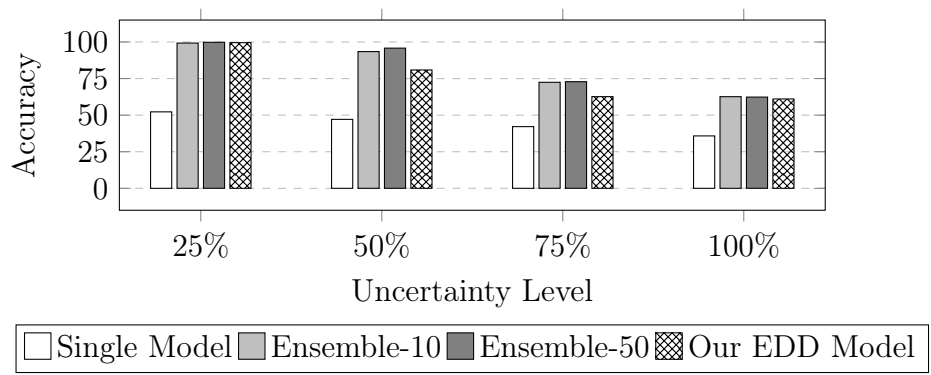

\begin{table}[H]
    \centering
    \def\arraystretch{1.25}
    \begin{tabular}{| c | c | c |}
        \hline
        \textbf{Model} & \textbf{Unperturbed data} & \textbf{Perturbed data $(\epsilon=0.1)$} \\
        \hline
        \hline
        \textbf{Single Model} & 0.8233 & 0.6653 \\
        \textbf{Ensemble 10} & 0.8800 & 0.8655 \\
        \textbf{Ensemble 50} & 0.8862 & 0.8807 \\
        \textbf{Ours} & 0.8598 & 0.7477 \\
        \hline
    \end{tabular}
    \vspace{0.5em}
    \caption{Comparison of AUC-ROC score of different models on unperturbed and highly perturbed UCI dataset.}
\end{table}
This is reflected in a slight decrease in the AUC-ROC for our distillation model in the adversarial case for our distillation model. We hypothesize that the comparatively small size of the UCI dataset, combined with the low diversity in the data, gives less chance for the distillation model to learn a robust representation of the ensemble's diversity. Further augmentation techniques may help to close the gap in these intermediate regions.
\subsubsection*{MotionSense}
Similarly to our experiments with HHAR, our experiments with our distillation model on Motion Sense gave significant increases in robustness at all levels of uncertainty (Figure 7) in line with the deployment of an ensemble, with the accuracy of our model at all uncertainty levels equivalent to an ensemble (Figure 8), and an AUC-ROC falling between 10-member and 50-member ensembles, even under significant adversarial perturbation (Table 5).
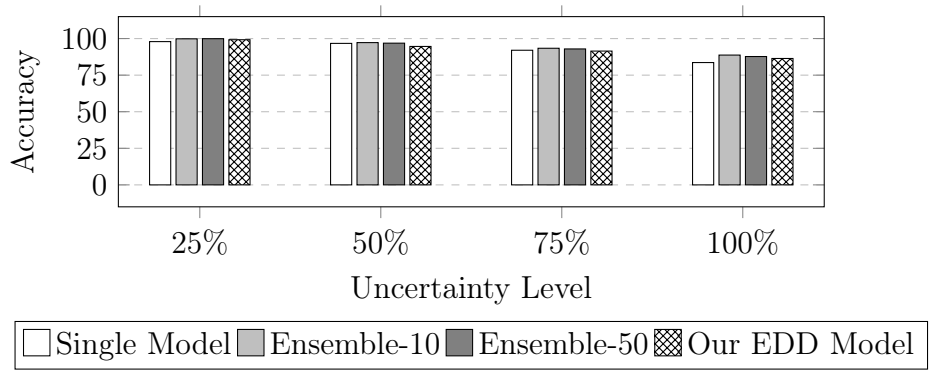
\begin{figure}[H]
    \centering
    \begin{tikzpicture}
    \begin{axis}[
        ybar,
        area legend,
        bar width=8pt,
        width=0.8\textwidth,
        height=0.3\textwidth,
        enlargelimits=0.15,
        legend style={
            at={(0.5,-0.6)},
            anchor=north,
            legend columns=-1
        },
        symbolic x coords={25\%,50\%,75\%,100\%},
        xtick=data,
        ylabel={Accuracy},
        ytick={0,25,50,75,100},
        xlabel={Uncertainty Level},
        ymin=0.0, ymax=100.0, 
        ymajorgrids=true,
        xmajorgrids=false,
        grid style=dashed,
        title={},legend image code/.code={
    \draw[#1,draw=black] (-5pt,-4pt) rectangle (5pt,6pt);
  },
    ]

    \addplot+[draw=black,fill=white]
        coordinates {(25\%,97.92) (50\%,96.71) (75\%,92.02) (100\%,83.60)};

    \addplot+[draw=black,fill=lightgray] 
        coordinates {(25\%,99.83) (50\%,97.24) (75\%,93.38) (100\%,88.72)};

    \addplot+[draw=black,fill=gray] 
        coordinates {(25\%,99.92) (50\%,96.86) (75\%,92.91) (100\%,87.73)};

    \addplot+[pattern=crosshatch,draw=black] 
        coordinates {(25\%,99.21) (50\%,94.62) (75\%,91.45) (100\%,86.40)};

    \legend{Single Model, Ensemble-10, Ensemble-50, Our EDD Model}

    \end{axis}
    \end{tikzpicture}
    \caption{Comparison of model accuracies on MotionSense with no adversarial perturbation at different uncertainty levels}
    \label{fig:model_accuracy}
\end{figure}
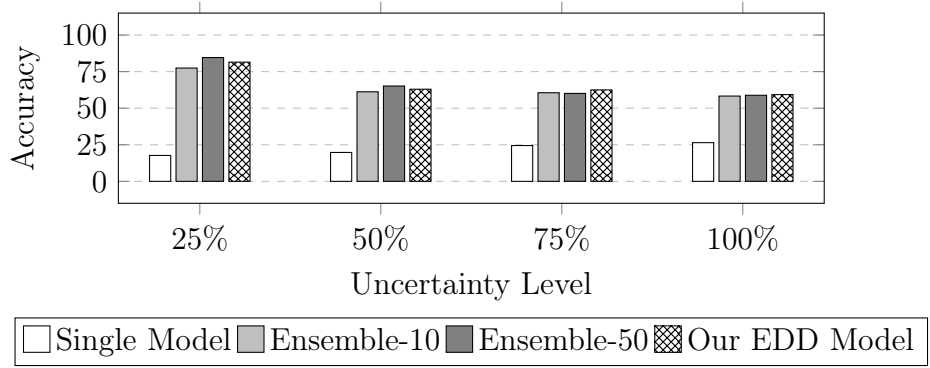
\begin{figure}[H]
    \centering
    \begin{tikzpicture}
    \begin{axis}[
        ybar,
        area legend,
        bar width=8pt,
        width=0.8\textwidth,
        height=0.3\textwidth,
        enlargelimits=0.15,
        legend style={
            at={(0.5,-0.6)},
            anchor=north,
            legend columns=-1
        },
        symbolic x coords={25\%,50\%,75\%,100\%},
        xtick=data,
        ylabel={Accuracy},
        ytick={0,25,50,75,100},
        xlabel={Uncertainty Level},
        ymin=0.0, ymax=100.0, 
        ymajorgrids=true,
        xmajorgrids=false,
        grid style=dashed,
        title={},legend image code/.code={
    \draw[#1,draw=black] (-5pt,-4pt) rectangle (5pt,6pt);
  },
    ]

    \addplot+[draw=black,fill=white]
        coordinates {(25\%,17.70) (50\%,19.76) (75\%,24.46) (100\%,26.44)};

    \addplot+[draw=black,fill=lightgray] 
        coordinates {(25\%,77.44) (50\%,61.21) (75\%,60.55) (100\%,58.29)};

    \addplot+[draw=black,fill=gray] 
        coordinates {(25\%,84.61) (50\%,65.16) (75\%,60.11) (100\%,58.84)};

    \addplot+[pattern=crosshatch,draw=black] 
        coordinates {(25\%,81.47) (50\%,62.98) (75\%,62.49) (100\%,59.28)};

    \legend{Single Model, Ensemble-10, Ensemble-50, Our EDD Model}

    \end{axis}
    \end{tikzpicture}
    \caption{Comparison of model accuracies on MotionSense with heavy adversarial perturbation ($\epsilon=0.1$) at different uncertainty levels}
    \label{fig:model_accuracy}
\end{figure}

\begin{table}[H]
    \centering
    \def\arraystretch{1.25}
    \begin{tabular}{| c | c | c |}
        \hline
        \textbf{Model} & \textbf{Unperturbed data} & \textbf{Perturbed data $(\epsilon=0.1)$} \\
        \hline
        \hline
        \textbf{Single Model} & 0.8277 & 0.3995 \\
        \textbf{Ensemble 10} & 0.8033 & 0.6058 \\
        \textbf{Ensemble 50} & 0.8111 & 0.6356 \\
        \textbf{Ours} & 0.7778 & 0.6217 \\
        \hline
    \end{tabular}
    \vspace{0.5em}
    \caption{Comparison of AUC-ROC score of different models on unperturbed and highly perturbed MotionSense dataset.}
\end{table}

\subsubsection*{PAMAP2}
For the PAMAP2 dataset, our results indicate that our distillation model demonstrates considerable robustness in accuracy and uncertainty estimation, particularly under conditions of significant adversarial perturbation (Figure 9). Under heavy adversarial perturbation our distillation model exhibits robust performance across all uncertainty levels, consistently surpassing the single-model baseline and matching, or exceeding, ensemble accuracy (Figure 10). These results emphasize its capability to reliably distinguish confident predictions from uncertain ones even under challenging adversarial conditions.
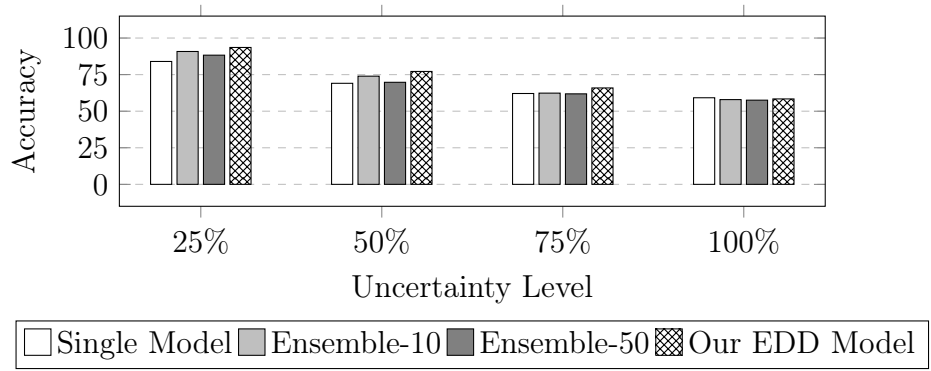
\begin{figure}[H]
    \centering
    \begin{tikzpicture}
    \begin{axis}[
        ybar,
        area legend,
        bar width=8pt,
        width=0.8\textwidth,
        height=0.3\textwidth,
        enlargelimits=0.15,
        legend style={
            at={(0.5,-0.6)},
            anchor=north,
            legend columns=-1
        },
        symbolic x coords={25\%,50\%,75\%,100\%},
        xtick=data,
        ylabel={Accuracy},
        ytick={0,25,50,75,100},
        xlabel={Uncertainty Level},
        ymin=0.0, ymax=100.0, 
        ymajorgrids=true,
        xmajorgrids=false,
        grid style=dashed,
        title={},legend image code/.code={
    \draw[#1,draw=black] (-5pt,-4pt) rectangle (5pt,6pt);
  },
    ]

    \addplot+[draw=black,fill=white]
        coordinates {(25\%,83.99) (50\%,69.06) (75\%,62.09) (100\%,59.18)};

    \addplot+[draw=black,fill=lightgray]
        coordinates {(25\%,90.81) (50\%,73.94) (75\%,62.35) (100\%,57.92)};

    \addplot+[draw=black,fill=gray] 
        coordinates {(25\%,88.29) (50\%,69.75) (75\%,61.85) (100\%,57.54)};
	
    \addplot+[pattern=crosshatch,draw=black] 
        coordinates {(25\%,93.56) (50\%,77.17) (75\%,65.89) (100\%,58.38)};

    \legend{Single Model, Ensemble-10, Ensemble-50, Our EDD Model}

    \end{axis}
    \end{tikzpicture}
    \caption{Comparison of model accuracies on PAMAP2 with no adversarial perturbation at different uncertainty levels}
    \label{fig:model_accuracy}
\end{figure}
\begin{figure}[H]
    \centering
    \begin{tikzpicture}
    \begin{axis}[
        ybar,
        area legend,
        bar width=8pt,
        width=0.8\textwidth,
        height=0.3\textwidth,
        enlargelimits=0.15,
        legend style={
            at={(0.5,-0.6)},
            anchor=north,
            legend columns=-1
        },
        symbolic x coords={25\%,50\%,75\%,100\%},
        xtick=data,
        ylabel={Accuracy},
        ytick={0,25,50,75,100},
        xlabel={Uncertainty Level},
        ymin=0.0, ymax=100.0, 
        ymajorgrids=true,
        xmajorgrids=false,
        grid style=dashed,
        title={},legend image code/.code={
    \draw[#1,draw=black] (-5pt,-4pt) rectangle (5pt,6pt);
  },
    ]
    
    \addplot+[draw=black,fill=white]
        coordinates {(25\%,26.90) (50\%,22.73) (75\%,25.52) (100\%,27.17)};

    \addplot+[draw=black,fill=lightgray]
        coordinates {(25\%,73.46) (50\%,56.02) (75\%,48.98) (100\%,46.48)};

    \addplot+[draw=black,fill=gray] 
        coordinates {(25\%,78.68) (50\%,58.38) (75\%,52.08) (100\%,49.86)};

    \addplot+[pattern=crosshatch,draw=black] 
        coordinates {(25\%,73.34) (50\%,55.83) (75\%,48.99) (100\%,45.07)};

    \legend{Single Model, Ensemble-10, Ensemble-50, Our EDD Model}

    \end{axis}
    \end{tikzpicture}
    \caption{Comparison of model accuracies on PAMAP2 with heavy adversarial perturbation ($\epsilon=0.1$) at different uncertainty levels}
    \label{fig:model_accuracy}
\end{figure}
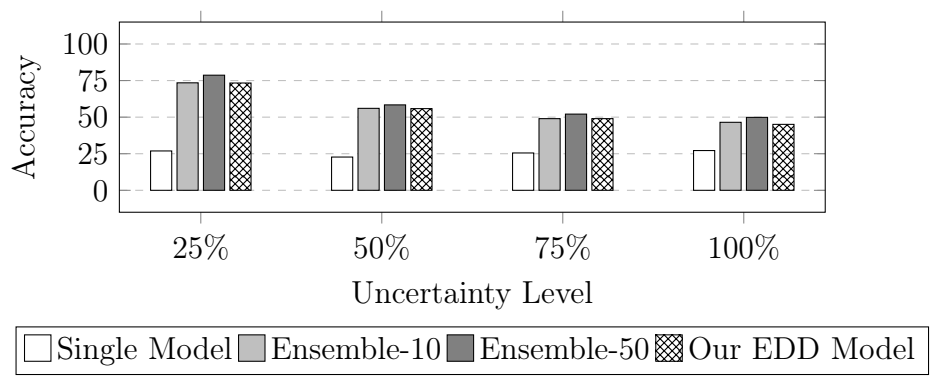

\begin{table}[H]
    \centering
    \def\arraystretch{1.25}
    \begin{tabular}{| c | c | c |}
        \hline
        \textbf{Model} & \textbf{Unperturbed data} & \textbf{Perturbed data $(\epsilon=0.1)$} \\
        \hline
        \hline
        \textbf{Single Model} & 0.6609 & 0.4587 \\
        \textbf{Ensemble 10} & 0.7254 & 0.6573 \\
        \textbf{Ensemble 50} & 0.6912 & 0.6457 \\
        \textbf{Ours} & 0.7840 & 0.6867 \\
        \hline
    \end{tabular}
    \vspace{0.5em}
    \caption{Comparison of AUC-ROC score of different models on unperturbed and highly perturbed PAMAP2 dataset.}
\end{table}
Our distillation model achieves the highest AUC-ROC scores of any model on both unperturbed and perturbed datasets, highlighting its superior capacity to effectively discriminate between accurate and inaccurate predictions (Table 6). These results underscore the effectiveness of our distillation approach in capturing the diversity of the ensembles, resulting in more reliable and robust uncertainty estimates.
\section{Conclusions and Future Work} \label{sec:conclusion}

In this paper, we presented a novel application of EDD within a self-supervised learning framework specifically tailored for HAR. Our approach addresses key challenges in HAR, such as the reliance on large labeled datasets and the need for robust and reliable models capable of providing meaningful uncertainty estimates.

By leveraging unlabeled data through self-supervised learning, we trained an ensemble of models on a pretext task and distilled their collective knowledge into a single prior network using EDD. This method captures both the predictive accuracy and the uncertainty estimates of the ensemble without increasing computational complexity at inference time. Our innovative data augmentation techniques, including the application of transformations and weighted combinations of samples, further enhanced the model's ability to generalize and quantify uncertainty effectively.

Our evaluations on publicly available HAR datasets demonstrated that the proposed method not only matches, but often surpasses the performance of traditional fully-supervised models trained on larger labeled datasets. Specifically, our approach achieved significant improvements in classification accuracy and the quality of uncertainty estimates, even under significant adversarial perturbation. This demonstrates the significant gains in robustness this approach yields.

In addition to its strong performance and uncertainty estimation capabilities, our approach offers significant computational advantages. By distilling the ensemble into a single prior network, we eliminate the inference-time overhead typically associated with UQ methods, ensuring that our method operates with the same efficiency as a standard single model. This makes it particularly well-suited for deployment on resource-constrained edge devices and wearables, where power and memory limitations are critical considerations. Despite this reduction in computational cost, our approach retains the predictive power and UQ of a full ensemble, making it a highly practical solution for real-world HAR applications.

The contributions of this work include:
\begin{itemize}
    \item Development of a self-supervised EDD framework for HAR that leverages unlabeled data to improve model performance.
    \item Introduction of innovative data augmentation techniques designed for time-series data in HAR, enhancing the model's robustness and UQ.
    \item Empirical validation showing that our method provides superior accuracy and robustness compared to baseline models and ensembles, without additional inference costs.
\end{itemize}

For future work, several avenues can be explored to build upon our findings:
\begin{itemize}
    \item Further investigation of data augmentation techniques for EDD in HAR and other domains.
    \item Further investigate the use of separable sources of uncertainty to improve outcomes in real-world applications, for example in active learning, or more sophisticated control schemes.
    \item Compare the domain adaptation capacity of semi-supervised models to our semi-supervised ensemble distillation model.
    \item Integrating inter-predictability frameworks that use uncertainty estimates to explain misclassifications or ambiguous predictions. This could be especially beneficial in safety-critical applications such as fall detection, where understanding the source of uncertainty can inform better risk management strategies
\end{itemize}

In conclusion, our self-supervised EDD framework offers a promising solution for advancing HAR by improving model accuracy, robustness and reliability without incurring additional costs associated with labeled data or increased computational demands. This work contributes to the broader effort of making deep learning models more accessible and trustworthy in applications where data scarcity and model uncertainty are significant concerns.

\section*{Acknowledgments}
This work was supported in part by Defence Science and Technology Group (DSTG).

\bibliographystyle{plain}
\bibliography{references}

\end{document}